\documentclass[journal]{IEEEtran}
\usepackage{graphicx}
\usepackage{float}
\usepackage{longtable}
\usepackage{tabularx}
\usepackage{color}

\usepackage[hidelinks]{hyperref}

\begin{document}

\title{Deep Learning based Computer Vision Methods for Complex Traffic Environments Perception: A Review}

\makeatletter
\newcommand{\linebreakand}{%
  \end{@IEEEauthorhalign}
  \hfill\mbox{}\par
  \mbox{}\hfill\begin{@IEEEauthorhalign}
}
\makeatother

\author{Talha Azfar~\IEEEmembership{Student Member, IEEE},
        Jinlong Li,
        Hongkai Yu, 
        Ruey Long Cheu~\IEEEmembership{Senior Member, IEEE},\\
        Yisheng Lv~\IEEEmembership{Senior Member, IEEE},
        and Ruimin Ke~\IEEEmembership{Member, IEEE}

        \thanks{Talha Azfar is with the Department
of Electrical and Computer Engineering, University of Texas at El Paso, El Paso, TX, 79968 USA, e-mail: tazfar@miners.utep.edu. Jinlong Li and Hongkai Yu are with Cleveland State University, Cleveland, OH 44115 USA, e-mails: j.li56@vikes.csuohio.edu and h.yu19@csuohio.edu. Ruey Long Cheu and Ruimin Ke are with the Department of Civil Engineering, University of Texas at El Paso, El Paso, TX, 79968 USA, e-mails: rcheu@utep.edu and rke@utep.edu. Yisheng Lv is with the Institute of Automation, Chinese Academy of Sciences, Beijing, 100190 China, e-mail: yisheng.lv@ia.ac.cn}
        \thanks{This work was supported in part by  NSF 2215388.}
        \thanks{* Corresponding author: Ruimin Ke, e-mail: rke@utep.edu}
}
\maketitle

\begin{abstract}
Computer vision applications in intelligent transportation systems (ITS) and autonomous driving (AD) have gravitated towards deep neural network architectures in recent years. While performance seems to be improving on benchmark datasets, many real-world challenges are yet to be adequately considered in research. This paper conducted an extensive literature review on the applications of computer vision in ITS and AD, and discusses challenges related to data, models, and complex urban environments. The data challenges are associated with the collection and labeling of training data and its relevance to real world conditions, bias inherent in datasets, the high volume of data needed to be processed, and privacy concerns. Deep learning (DL) models are commonly too complex for real-time processing on embedded hardware, lack explainability and generalizability, and are hard to test in real-world settings. Complex urban traffic environments have irregular lighting and occlusions, and surveillance cameras can be mounted at a variety of angles, gather dirt, shake in the wind, while the traffic conditions are highly heterogeneous, with violation of rules and complex interactions in crowded scenarios. Some representative applications that suffer from these problems are traffic flow estimation, congestion detection, autonomous driving perception, vehicle interaction, and edge computing for practical deployment. The possible ways of dealing with the challenges are also explored while prioritizing practical deployment.
\end{abstract}


\IEEEpeerreviewmaketitle

\section{Introduction}

\begin{figure} 
    \begin{centering}
        \includegraphics[width=1\columnwidth]{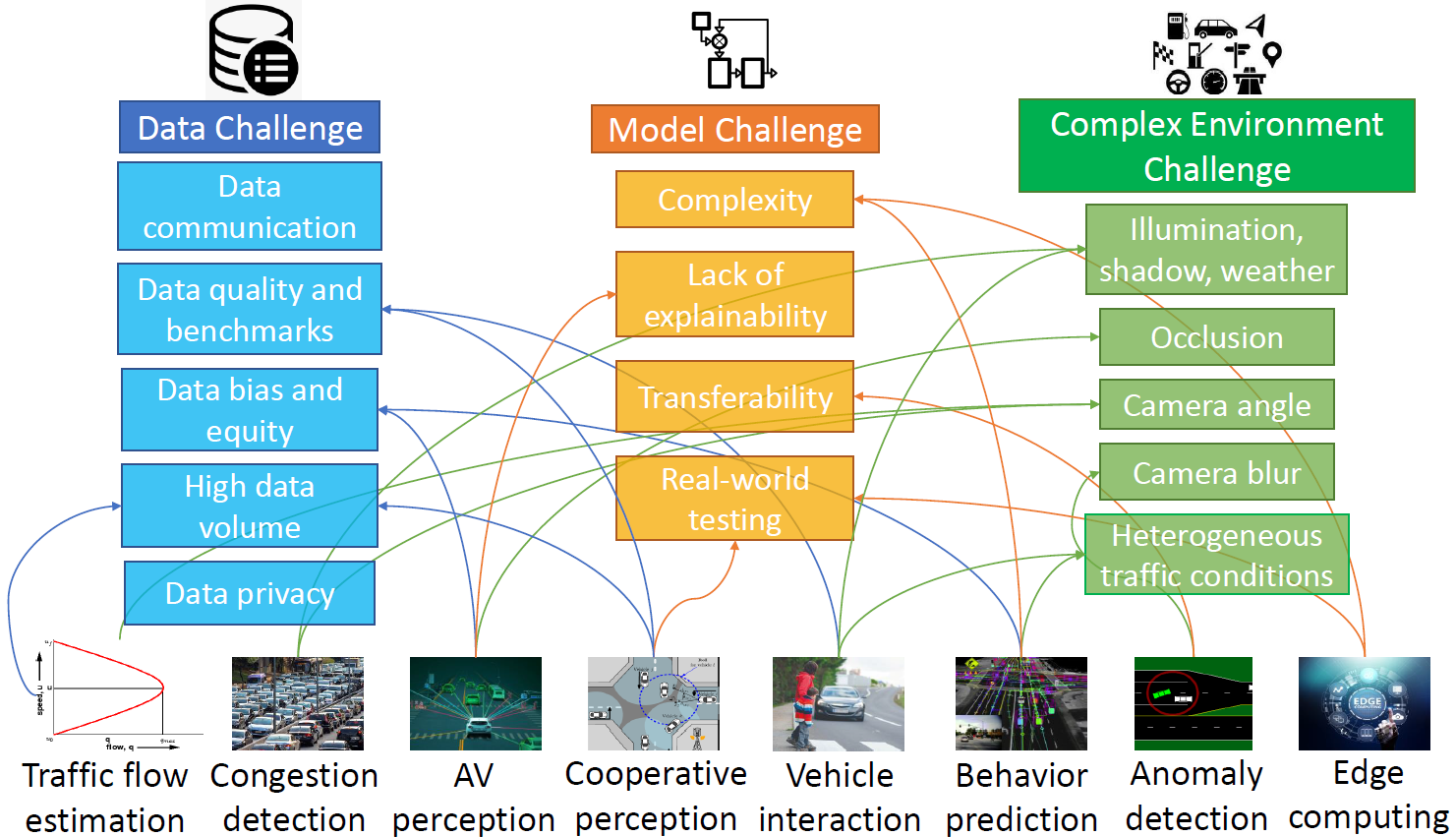}
    \par\end{centering}
    \caption{Applications of deep learning based computer vision methods in transportation and associated challenges in real world deployment.}
    \label{fig:introduction}
\end{figure}

Video cameras have been used to monitor traffic and provide valuable information to traffic management center (TMC) operators. The manual process of having TMC operators observe numerous video screens has given way to automated and semi-automated computer vision approaches for faster processing and response times with some humans in the loop to interpret and verify the data. Artificial neural networks are being increasingly used in computer vision in ITS and autonomous driving (AD) applications, showing benefits in traffic monitoring, traffic flow estimation, incident detection, etc. However, the use of deep neural networks (DNN) brings some issues and concerns that should be studied in further detail as they need to be more accurate, more reliable, and practical enough for large-scale deployment as part of ITS infrastructure or in autonomous vehicles. 

Deep learning (DL) refers to machine learning architectures of multiple layers such as large (deep) neural networks spanning many layers in a variety of configurations. Advances in computational hardware and algorithmic efficiency have made DL popular in every field that deals with big data. DNNs have been applied to solve computer vision problems such as object detection, classification, motion tracking, and prediction \cite{o2019deep}. ITS and AD researchers have adapted these models, training them for specific use cases with specially curated datasets and benchmarks such as KITTI \cite{kitti2012} and AI City \cite{naphade2019aicity}. 

Automated analysis of traffic surveillance videos in ITS systems is important for incident and congestion management, while perception in autonomous vehicles is critical for vehicle control and navigation. Computer vision algorithms used for these purposes must therefore be scrutinized in detail and all of the possible problems should be addressed in advance before real-world deployment. In the course of this literature review, a number of recurrent issues were discovered related to the data, models, and complex urban environments which are detailed in this paper. Large quantities of data are necessary for training DNNs and evaluating their performance, but this poses issues such as over-representation of common events or classes, time and effort required to label and select data, and a lack of consistent benchmarks for a fair evaluation. Complex DL models can be trained to infer more accurately but this comes at the cost of efficiency, lack of explainability, and difficulty in adapting the solution to diverse or unseen use cases (not present in the training set). Real-world uncertainties involved in complex urban environments like shadow, lighting, and occlusion are common issues, while variable surveillance camera angles and heterogeneous traffic conditions present further challenges to DNNs even after training on these conditions.


While these issues have been mentioned in some of the literature, only a few approaches have been developed to address them, and even fewer real-world implementation examples were found. Computer vision in transportation is a very active research field, and over 200 papers were selected and reviewed for this article. Figure \ref{fig:introduction} gives an overview of the applications and challenges for quick reference, while Table \ref{table1} summarizes the methods used in each application and associated challenges. The following sections (\ref{Sec:Data}, \ref{Sec:Model}, \ref{Sec:Complex}) discuss the specific challenges for data, models, and complex traffic environments. A number of representative applications and solutions to meet the challenges are explained in section \ref{Sec:Applications}. This is followed by section \ref{Sec:Future}, a collection of future directions that research in this area should take. Finally, section \ref{Sec:Conclusions} presents some concluding remarks.

The contributions of this paper are:
\begin{itemize}
    \item Classification of common challenges faced by computer vision DL methods in complex traffic environments.
    \item A review of DL models used for some representative computer vision applications susceptible to the challenges.
    \item Specific techniques already being used to mitigate the challenges.
    \item Future directions of research to improve DL models for real world complex traffic environments. 
\end{itemize}

\section{Data challenges}\label{Sec:Data}

\subsection{Data communication}
Data communication, while not considered in most ITS and AV computer vision studies in the lab, is critical in practical applications. Individual-camera-based deep learning tasks in practice commonly require data communication between the camera and the cloud server at TMC. Video data is in rich volume, which can cause potential data communication issues, such as transmission delay and package loss. In a cooperative camera-sensing environment, there are not only data communications with the server but also among different sensors. Therefore, two additional issues are multi-sensor calibration and data synchronization. 

Calibration in a cooperative environment aims to determine the perspective transformation between sensors to be able to merge acquired data from several views at a given frame \cite{caillot2022survey}. This task is quite challenging in a multi-user environment, because the transformation matrix between sensors constantly changes as the vehicles move.
In a cooperative context, calibration relies on the synchronization of the elements in a background image to determine the transformation between static or mobile sensors~\cite{yang2021synch}. There are multiple sources of desynchronization, such as an offset between the clocks or variable communication delays. Although clocks may be synchronized, it is difficult to ensure the data acquisitions are triggered at the same moment which adds uncertainty towards merging the acquired data. Similarly, different sampling rates require interpolation between acquired or predicted data, also adding uncertainty.

\begin{figure} 
    \begin{centering}
        \includegraphics[width=1\columnwidth]{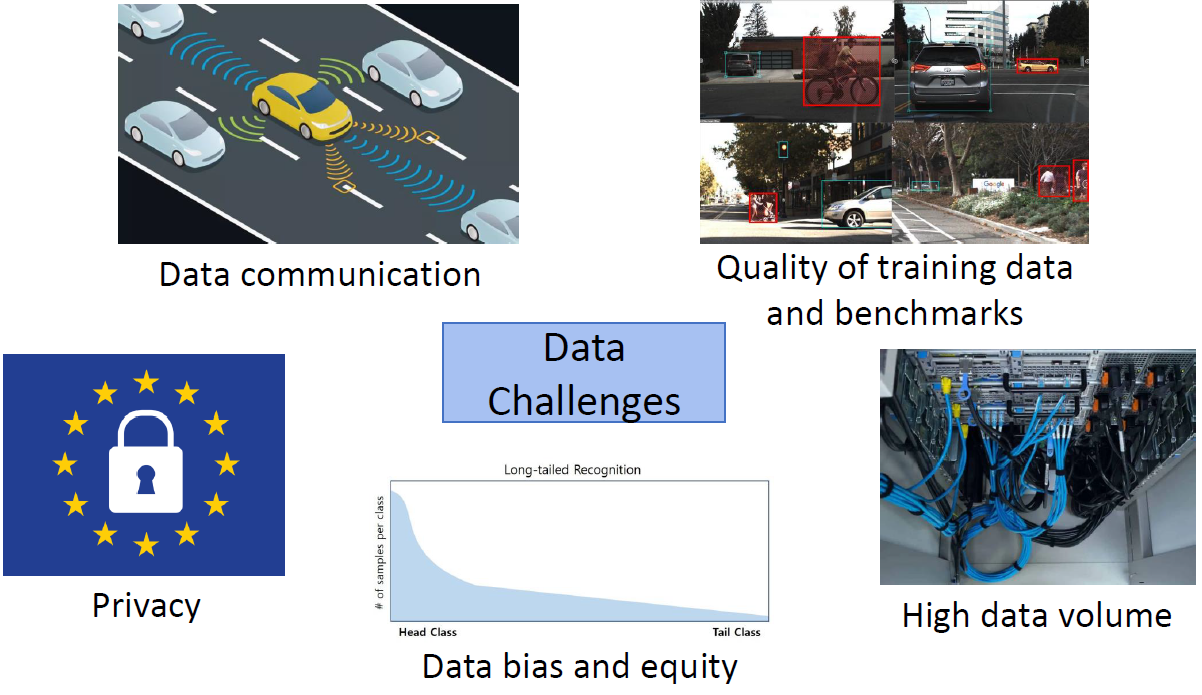}
    \par\end{centering}
    \caption{Illustration of representative challenges associated with data in computer vision applications for transportation.}
    \label{fig:datachallenge}
\end{figure}

\subsection{Quality of training data and benchmarks}

Traffic cameras are widely deployed on roadways and vehicles \cite{ke2020real}. TMCs at DOTs and cities constantly collect network-wide traffic camera data, which are valuable for various ITS applications, such as event recognition and vehicle detection. However, labeled training data is much less common than unlabeled data \cite{halevy2009unreasonable,luo2018mio}. The lack of annotated datasets for many applications is slowly being overcome with synthetic data, as graphical fidelity and simulated physics have become more and more realistic. For example, ground truth 3D information in \cite{Hu_2019_ICCV} needs high accuracy during training for monocular 3D detection and tracking, so video game data was used. In addition to realistic appearance, simulated scenarios do not need to be manually labeled as the labels are already generated by the simulation, and can support a wide variety of illuminations, viewpoints, and vehicle behaviors \cite{yao2020simulating}. The 2020 AI City challenge for vehicle re-identification winner utilized a hybrid dataset to significantly improve the performance \cite{zheng2020going} by generating examples from real-world data and adding other simulated views and environments. However, if using synthetic data, additional learning procedures, e.g., domain adaptation, are still needed for real-world applications. Low-fidelity simulated data were used to train a real-world object detector with domain randomization transfer learning \cite{tobin2017domain}.  

 The lack of good quality crash and near-crash data is often cited as a practical limitation \cite{Taccari2018crash}. More crash data will update the attention guidance in AD, allowing it to capture long-term crash characteristics, thereby improving crash risk estimation \cite{YuLiCrash2021}. There is also a lack of representation in the literature regarding bicycles as the ego vehicle as mentioned in \cite{ibrahim2021cyclingnet}.
 A near-miss incident database was developed in \cite{kataoka2018nearmiss} to compensate for the unavailability, however, it is private because of copyright issues.
 A review of vehicle behavior prediction methods \cite{Mozaffari2022} discusses the lack of a benchmark for evaluating existing studies, preventing a fair comparison of different DL techniques, or classical methods like Bayesian or Markov decision process. It also highlights that faulty or limited sensors, constrained computational resources, and generalizability to any driving scenario are current barriers to practical deployment and represent a significant research gap. Some of these issues can be addressed by sensor fusion, internet of vehicles (IoV), and edge computing \cite{wang2020sensorfusion}.

\subsection{Data bias}

Although current vehicle detection algorithms perform well on balanced datasets, they suffer from performance degradation on tail classes when facing imbalanced datasets. In real-world scenarios, data tends to obey the Zipfian distribution~\cite{reed2001pareto} where a large number of tail categories have few samples. In long-tail datasets, a few head classes (frequent classes) contribute most of the training samples, while tail classes (rare classes) are underrepresented.
Most DL models trained with such data minimize empirical risk on long-tail training data and are biased towards head categories since they contribute most of the training data~\cite{wang2022c2am, fu2021let}.
Some methods like data re-sampling~\cite{mahajan2018exploring} and loss re-weighting~\cite{wang2021seesaw}, can compensate the underrepresented classes. However, they need to partition the categories into several groups based on their category frequency prior. Such hard division between head and tail classes brings two problems: training inconsistency between adjacent categories and lack of discriminative power for rare categories~\cite{wang2021adaptive}.
 
 An object detection study focusing on construction vehicles found that training a deep model with huge general training dataset did not perform as well as a smaller model trained specifically on construction vehicles \cite{arabi2020deep}. Another model based on YOLOv2 for vehicle size estimation performed well on commonly seen sizes but varied considerably with uncommon sizes \cite{wu2019multiscale}. The dataset used by \cite{garcia2021classimbalance} for autonomous vehicle object detection had severe class imbalance with only 1\% cyclists represented. A number of weight based learning strategies were employed to address this, giving higher weight to underrepresented classes, showing significant improvements.
 
General object detectors can be improved using transfer learning with the underrepresented data for task specific performance benefits \cite{zhao2019objreview}. Also, it is noted in \cite{ras2018expl} that model bias may not always be apparent from just the training set, and explanability methods are needed to address the problem.

\subsection{High data volume} 
 Visual data is composed of over 90\% of the Internet traffic and video transmission, computation, and storage pose increasing challenges in ITS and AV fields \cite{ke2020real}. The high volume of traffic and vehicular-based video data from the roadside and onboard sensors via the traffic camera network or the Internet of Vehicles (IoV) network poses computational and bandwidth bottlenecks that cannot be solved by using more powerful equipment \cite{xu2018bigdata}. As many applications in connected or autonomous vehicles rely on DL, vehicle-cloud architecture is emerging as an effective distributed computing technique \cite{wang2011cloudvehicle}. With the integration of Road Side Units (RSU), these edge nodes can process faster and provide low communication latency.

\subsection{Security and Privacy}

 Privacy concerns are an important human factor that cannot be overlooked in the design and operation of ITS applications~\cite{fries2012meeting}.
 Observing and tracking the massive amounts of the pedestrian and vehicle information causes security and privacy concerns in ITS environments. For example, UAVs are capable of collecting traffic data (through onboard video cameras). However, privacy concerns restrict them from being a regular part of the ITS sensor network ~\cite{khan2021provable}.
 Video surveillance systems constantly collect human faces and license plates. Personal privacy is exchanged for security or safety services provided by the surveillance \cite{martinez2013pursuit}. 
 Systems deployed in practice might need to de-identify faces and license plates in real-time if raw video data is being sent or stored \cite{martinez2012towards}. Any processing would ideally be done on the local edge unit, limiting the propagation of private information. Full anonymity is difficult to guarantee, for example, an uncommon model car with distinctive paint pattern can be traced to its owner by correlating with other information, even with a blurred license plate.

\begin{figure}[h] 
    \begin{centering}
        \includegraphics[width=1\columnwidth]{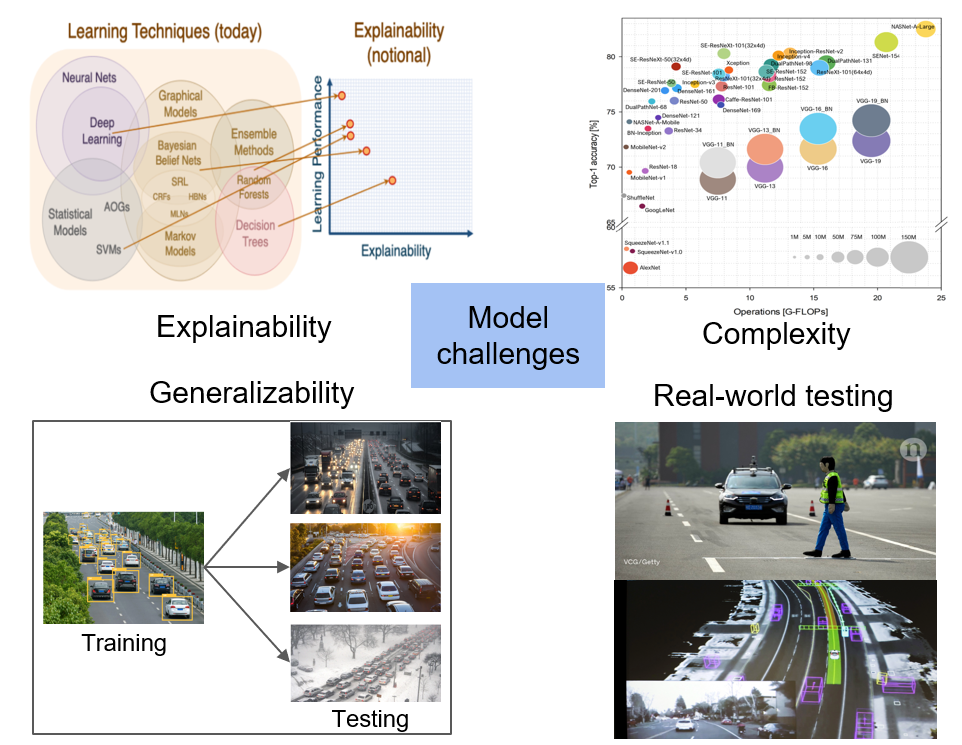}
    \par\end{centering}
    \caption{Illustration of representative model challenges. Some demo images are adopted from~\cite{bianco2018benchmark, bornstein2016artificial}.}
    \label{modelchallenge}
\end{figure}

\section{Model challenges}\label{Sec:Model}

\subsection{Complexity}

DL computer vision models have a high complexity with respect to neural network structures and training procedures. Many DL models are designed to run on high-performance cloud centers or AI workstations, and a good model requires weeks or months of training as well as high power consumption driven by modern Graphical Processing Units (GPUs) or Tensor Processing Units (TPUs).

In ITS and AV field applications, a lot of them have a requirement for real-time or near real-time operations \cite{ke2020real} for the sake of functionality and traffic safety. The DL model complexity adds a high cost to the training and inference in real-time applications; particularly, the trend of ITS and AV is towards large-scale on-device processing closer to where the traffic data is generated, e.g., crowdsensing. Three popular embedded devices are compared in \cite{arabi2020deep}, with the Nvidia Jetson Nano yielding the highest inference efficiency but it is too little computation power for complex applications. 

Real-time applications usually make some modifications like resizing video to lower resolution or model quantization and pruning which can lead to loss of performance. The model complexity of the state-of-the-art DL methods needs to be reduced in many practical applications in order to meet the efficiency and accuracy requirements. For example, multi-scale deformable attention has been used with vision transformer neural networks in object detection for high performance and fast convergence leading to faster training and inference \cite{zhu2021deformable}.

\subsection{Lack of explainability}

DNNs are largely seen as black boxes with many layers of processing, the working of which can be examined using statistics, but the learned internal representations of the network are based on millions or billions of parameters, making analysis extremely difficult \cite{ras2018expl}. This means that the behavior is essentially unpredictable, and very little explanation can be given of the decisions. It also makes system validation and verification impossible for critical use cases like autonomous driving \cite{samek2017expl}.  

The common assumption that a complex black box is necessary for good performance is being challenged \cite{rudin2019stop}. Recent research is attempting to make DNNs more explainable. A visualization tool for vision transformers is presented in \cite{aflalo2022vl}, which can be used to see the inner mechanisms, such as hidden parameters, and gain insight into specific parts of the input that influenced the predictions. A framework for safety, explainability, and regulations for autonomous driving was evaluated in post-accident scenarios \cite{atakishiyev2021expl}. The results showed many benefits including transparency and debugging. A convolutional neural network (CNN) based architecture is proposed to detect action-inducing objects for autonomous vehicles, while also providing explanations for the actions \cite{xu2020expl}. 

\subsection{Transferability and generalizability}

Generalization to out-of-distribution data is natural to humans yet challenging for machines because most learning algorithms strongly rely on the independent and identically distributed (i.i.d.) assumption training at testing data, which is often violated in practice due to domain shift.
Domain generalization aims to generalize models to new domains without knowledge about the target distribution during training. Different methods have been proposed for learning generalizable and transferable representations~\cite{dou2019domain}.

Most existing approaches belong to the category of domain alignment, where the main idea is to minimize the difference among source domains for learning domain-invariant representations. Features that are invariant to the source domain shift should also be robust to any unseen target domain shift. Data augmentation has been a common practice to regularize the training of machine learning models to avoid overfitting and improve generalization~\cite{lecun2015deep}, which is particularly important for over-parameterized DNNs.

Visual attention in CNNs can be used to highlight the regions of the image involved in a decision, with causal filtering to find the most relevant parts \cite{Kim2018book}. The importance of individual pixels is estimated in \cite{petsiuk2018rise} by using randomly masked versions of images and comparing the output predictions. This approach does not apply to spatio-temporal methods or those that consider relationships between objects in complex environments. 

\subsection{Real-world testing}

In general, DL methods have been shown to be prone to underspecification, a problem that appears regardless of model type or application. Among other domains, underspecification in computer vision is analyzed in \cite{underspecification2020}, specifically for DL models such as the commonly used ResNet-50 and a scaled-up transfer learning image classification model, Big Transfer (BiT) \cite{BiT2019}. It is shown that while benchmark scores improved with more model complexity and training data, testing with real-world distortions results in poor and highly varied performance that depends strongly on the random seeds used to initialize training.

Practical systems need to be efficient in terms of memory and computation for real-time processing on a variety of low-cost hardware \cite{BAI2021expl}. Some approaches towards efficient and low-cost computation include parameter pruning, network quantization, low-rank factorization, and model distillation. Approaches like \cite{cuitrajectory2019} are efficient and capable of real-time trajectory prediction but are not end-to-end because they assume the prior existence of an object tracking system to estimate the states of surrounding vehicles. 

Vulnerable road users (VRU) such as pedestrians and bicyclists present a unique problem, since they can change their direction and speeds very rapidly, and interact with the traffic environment differently than vehicles \cite{saleh2017VRU}.

 Some of the major barriers to the practical deployment of computer vision models in ITS are the heterogeneity of data sources and software, sensor hardware failure, and extreme or unusual sensing cases \cite{Zhou2021edge}. Furthermore, recent frameworks such as those based on edge computing directly expose the wireless communication signals of a multitude of heterogeneous devices with various security implementations, creating an ever-increasing potential attack surface for malicious actors \cite{contreras2018iov}. Deep learning models have been developed to detect these attacks, however real-time application and online learning are still areas of active research \cite{chen2019edgesecure}. 
 
 IoV faces fundamental practical issues arising from the fact that moving vehicles will present highly variable processing requirements on the edge nodes, while each vehicle can also have many concurrent edge and cloud related applications running, along with harsh wireless communication environments \cite{zhangEdge2020}. Other challenges related to edge computing for autonomous vehicles include cooperative sensing, cooperative decisions, and cybersecurity \cite{liu2019edge}. Attackers can use lasers and bright infrared light to interfere with cameras and LiDAR, change traffic signage, and replay attacks over the communication channel.
 A visual depiction of model challenges can be seen in Figure \ref{modelchallenge}.

\section{Complex Traffic Environments}\label{Sec:Complex}

\begin{figure} 
    \begin{centering}
        \includegraphics[width=1\columnwidth]{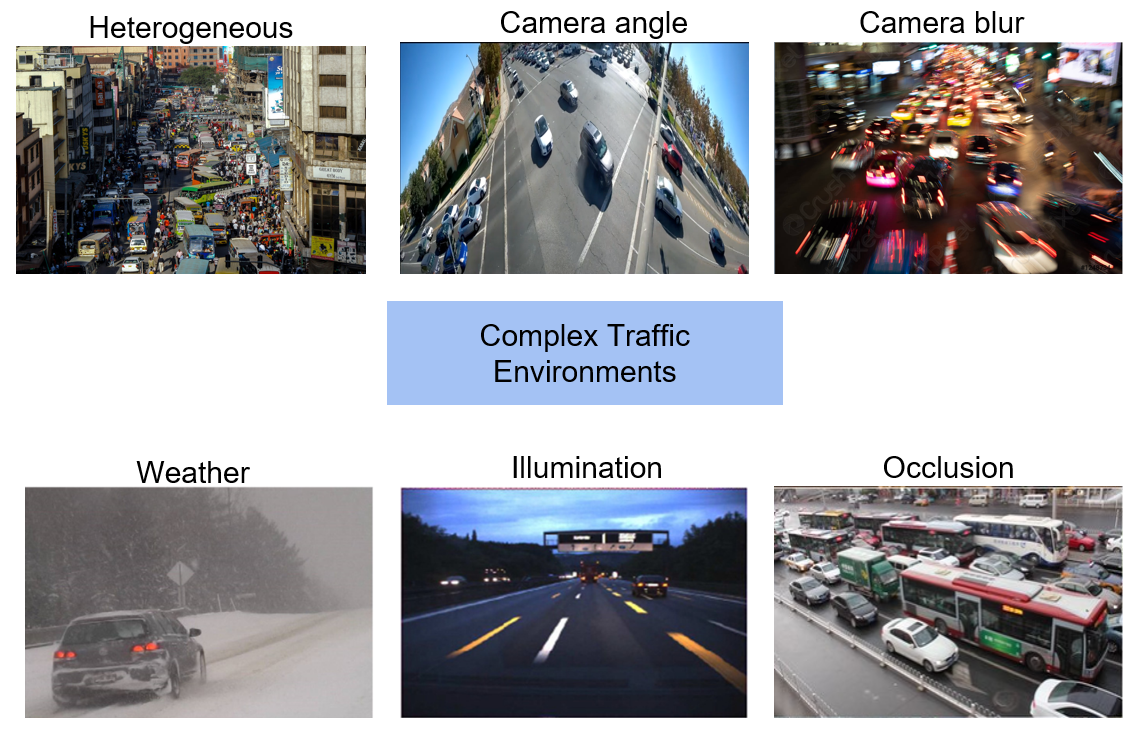}
    \par\end{centering}
    \caption{Illustration of representative scenarios in complex traffic environments. Some demo images are adopted  from~\cite{yang2018vehicle}.}
    \label{complexenvironment}
\end{figure} 

\subsection{Shadow, lighting, weather}

Situations like shadows, adverse weather, similarity between background and foreground, strong or insufficient illumination in the real world are cited as common issues~\cite{lin2019moving, song2020deep}. The appearance of camera images is known to be affected by adverse weather conditions, such as heavy fog, sleeting rain, snowstorms, and dust storms~\cite{hassaballah2020vehicle}.

A real-time crash detection method in \cite{jiansheng2014vision} utilizes foreground extraction using Gaussian Mixture Model, then tracks vehicles using a mean shift algorithm. The position, speed, and acceleration o the vehicles are passed through threshold functions to determine the detection of a crash. While computationally efficient, such methods suffer significantly in the presence of noise, complex traffic environment, and change in weather.

In harsh weather conditions, vehicles captured by traffic surveillance cameras exhibit issues such as underexposure, blurring, and partial occlusion. At the same time, raindrops, and snowflakes that appear in traffic scenes add difficulty for the algorithm to extract vehicle targets~\cite{yang2018vehicle}. At night, or in tunnels with vehicles driving towards the camera, the scene may be masked completely because of the high beam glare~\cite{sonnleitner2020traffic}. 

\subsection{Occlusion}

Occlusion is one of the most challenging issues, where a target object is only partially available to the camera or sensor due to obstruction by another foreground object. Occlusion exists in various forms ranging from partial occlusion to heavy occlusion~\cite{gilroy2019overcoming}. In AD, target objects can be occluded by static objects such as buildings and lampposts. Dynamic objects such as moving vehicles or other road users may occlude one another, such as in crowds. Occlusion is also a common issue in object tracking \cite{nowosielski2016traj} because once the tracked vehicle disappears from view and reappears, it is considered a different vehicle causing tracking and trajectory information to be inaccurate. 

\subsection{Camera angle}

In the applications of transportation infrastructure, the diversity of surveillance cameras and their viewing angles pose challenges to DL methods trained on limited types of camera views \cite{buch2011review}\cite{santhosh2020anomalySurvey}. 
While the algorithm in \cite{Albiol2011} is computationally efficient and can work in varying lighting conditions and traffic density scenarios, lower-pitch camera views and road marking corners can introduce significant errors. The model in \cite{armstrongCVPR2021} can identify anomalies near the camera, including their start and end times, but is not accurate for anomalies in the distance since the vehicles occupy only a few pixels. 

An earlier survey on anomaly detection from surveillance video concluded that illumination, camera angle, heterogeneous objects, and a lack of real-world datasets are the major challenges \cite{santhosh2020anomalySurvey}. The methods used for sparse and dense traffic conditions are different and lack generalizability. Matching objects in different views is another major problem in a multi-view vision scene, as multi-view ITS applications need to process data across the different images captured by different cameras at the same time~\cite{xie2021deep}.

\subsection{Camera blur and degraded images}
Surveillance cameras are subject to weather elements. Water, dust, and particulate matter can accumulate on the lens causing image quality degradation. Strong wind can cause a camera to shake, resulting in motion blur in the whole image. Front-facing cameras on autonomous vehicles also face this, as insects can smash onto the glass, causing blind spots in the camera's field of view.
Specifically, object detection and segmentation algorithms suffer greatly \cite{vasiljevic2017examining}, and unless preparations are made in the model, false detections can cause serious safety issues in AD and miss important events in surveillance applications. Some approaches to address this include using degraded images for training, image restoration preprocessing, and fine-tuning pre-trained networks to learn from degraded images. For example, Dense-Gram networks are used in \cite{guo2019degraded} which improve image segmentation performance in degraded images.

\subsection{Heterogeneous, urban traffic conditions}
Dense urban traffic scenarios are full of complex visual elements, not only in quantity but also in the variety of different vehicles and their interactions, as shown in Figure \ref{complexenvironment}. The presence of cars, buses, bicycles, and pedestrians in the same intersection is a significant problem for autonomous navigation and trajectory computation \cite{ma2018autorvo}. The different sizes, turning radii, speeds, and driver behaviors are further compounded by the interactions between these road users.
From a DL perspective, it is easy to find videos of heterogeneous urban traffic, but labeling for ground truth is very time-consuming. Simulation software usually cannot capture the complex dynamics of such scenarios, especially the traffic rule-breaking behaviors seen in dense urban centers. In fact, a specific dataset was created to represent these behaviors in \cite{traphic2019}. A simulator for unregulated dense traffic was created in \cite{cai2020summit} which is useful for autonomous driving perception and control but does not represent the trajectory and interactions of real-world road users.

\section{Applications}\label{Sec:Applications}

\subsection{Traffic flow estimation}
\subsubsection{Models and algorithms}
Traffic flow variables include traffic volume, density, speed, and queue length. The algorithms and models to detect and track objects to estimate traffic flow variables from videos may be classified into one-stage and two-stage methods.
In one-stage methods, the variables are estimated from detection results and there is no further classification and location optimization, for example: 1) YOLOv3 + DeepSORT tracker~\cite{hong2020traffic}; 2) YOLOv2 + spatial pyramid pooling~\cite{kim2019multi}; 3) AlexNet + optical flow + Gaussian mixture model~\cite{ke2018multi}; 4) CNN + optical flow based on UAV video~\cite{ke2018real}; 5) SSD (single shot detection) based on UAV video~\cite{tang2017arbitrary}.

Two-stage methods first generate region proposals that contain all potential targets in the input images and then conduct classification and location optimization. Examples of two-stage methods are: 1) Faster R-CNN + SORT tracker~\cite{fedorov2019traffic}; 2) Faster R-CNN~\cite{peppa2018urban, mhalla2018embedded}; 3) Faster R-CNN based on UAV video~\cite{peppa2021towards, brkic2020analytical}.

\subsubsection{Current methods to overcome challenge}
A DL method at the edge of the ITS that performs real-time vehicle detection, tracking, and counting in traffic surveillance video has been proposed in \cite{chen2020edge}. The neural network detects individual vehicles at the single-frame level by capturing appearance features with the YOLOv3 object-detection method, deployed on edge devices to minimize bandwidth and power consumption. A vehicle detection and tracking approach in adverse weather conditions that achieves the best trade-off between accuracy and detection speed in various traffic environments is discussed in \cite{hassaballah2020vehicle}. Also, a novel dataset called DAWN is introduced for vehicle detection and tracking in adverse weather conditions like heavy fog, rain, snow, and sandstorms, in order to make training less biased.

\subsection{Traffic congestion detection}
\subsubsection{Models and algorithms}
The methods that detect traffic congestion based on computer vision may also be divided into one-stage methods and multi-step methods. The one-stage methods identify vehicles from the video images and directly perform traffic congestion detection. Among the one-stage methods are: (1) AlexNet and YOLO~\cite{chakraborty2018traffic} to distinguish congestion and non-congestion, (2) AlexNet and VGGNet~\cite{wang2020towards} which classify `jam' and `no jam'; and (3) YOLO and Mask R-CNN~\cite{impedovo2019vehicular} recognize light, medium, and heavy congestion (identifying the number of vehicles in each frame and then classify). The multi-step methods first apply traffic flow estimation models to measure traffic variables and then use the traffic flow variables to infer congestion. Examples of two-stage traffic congestion detection models are: (1) YOLOv3~\cite{rashmi2020vehicle} and YOLOv4~\cite{sonnleitner2020traffic} for vehicle detection and counting, (2) counting vehicles using Faster R-CNN~\cite{gao2021novel} and applying regression for traffic congestion. Beside these, the traffic congestion can be evaluated by the traffic flow detection algorithms~\cite{kumar2021applications} using vehicle detection and tracking.

\subsubsection{Current methods to overcome challenge}
Congestion detection performance can be improved using multiple sensors based solutions including radar, lasers, and sensor fusion since it is hard to achieve ideal performance and accuracy using a single sensor in real-world scenarios. There is a wide use of decision-making algorithms for processing fusion data acquired from multiple sensors~\cite{muhammad2020deep}. A CNN-based model trained with bad weather condition datasets can improve the detection performance~\cite{sharma2022deep}, while generative adversarial network (GAN) based Style Transfer methods have also been applied~\cite{lin2020gan}\cite{li2021domain}. These approaches help to minimize the model challenges related to generalizability, which in turn improves real-world performance.

\subsection{Autonomous driving perception: detection}
\subsubsection{Models and algorithms}

Common detection tasks that assist in AD are categorized into traffic sign detection, traffic signal detection, road/lane detection, pedestrian detection, and vehicle detection.

There are two tasks in a typical traffic sign recognition system: finding the locations and sizes of traffic signs in natural scene images (traffic sign detection) and classifying the traffic signs into their specific sub-classes (traffic sign classification)~\cite{yang2015towards}.
An improved Sparse R-CNN was used for traffic sign detection in~\cite{cao2021traffic}, while an efficient algorithm based on YOLOv3 model for traffic sign detection was implemented in~\cite{wan2021efficient}. SegU-Net, formed by merging the state-of-the-art segmentation architectures SegNet and U-Net to detect traffic signs from video sequences has been proposed \cite{kamal2019automatic}.
Several adaptations to Mask R-CNN were tested in ~\cite{tabernik2019deep} for detection and recognition with end-to-end learning on the domain of traffic signs. They also proposed a data augmentation technique based on the distribution of geometric and appearance distortions.

A method that uses an encoder-decoder DNN with focal regression loss to detect small traffic signals is proposed in \cite{lee2019accurate}. 
It is shown in~\cite{kim2018efficient} that Faster R-CNN with Inception-Resnet-v2 model is more suitable for traffic light detection than others.
A practical traffic light detection system in~\cite{ouyang2019deep} combines CNN classifier model and heuristic region of interest (ROI) candidate detection on self-driving hardware platform Nvidia Jetpack Tx1/2 that can handle high-resolution images.
The recognition accuracy and processing speed are improved by combining detection and tracking in~\cite{wang2021simultaneous} to enhance the practicality of the traffic signal recognition system in autonomous vehicles using CNN and integrated channel feature tracking to determine the coordinates and color for traffic lights.

Lane detection aims to identify the left and right lane boundaries from a processed image and apply an algorithm to track the road ahead. A novel hybrid neural network combining CNN and recurrent neural network (RNN) for robust lane detection in driving scenes has been proposed~\cite{zou2019robust}. Features on each frame of the input video were first abstracted by a CNN encoder and the sequential encoded features were processed by a ConvLSTM. The outputs were fed into the CNN decoder for information reconstruction and lane prediction.
Another lane detection method is an anchor-based single-stage lane detection model called LaneATT~\cite{tabelini2021keep}. It uses a feature pooling method with a relatively lightweight backbone CNN, while maintaining high accuracy. A novel anchor-based attention mechanism to aggregate global information was also proposed. 
A new method to impose structure on badly posed semantic segmentation problems is proposed in \cite{ghafoorian2018gan} by using a generative adversarial network architecture with a discriminator that is trained on both predictions and labels at the same time.

A two-stage detector SDS-RCNN~\cite{brazil2017illuminating} jointly learned pedestrian detection and bounding-box aware semantic segmentation, thus encouraging model learning on pedestrian regions.
RPN+BF~\cite{zhang2016faster} used a boosted forest to replace second-stage learning and leveraged hard mining for proposals. However, involving such downstream classifiers could bring more training complexity. AR-Ped~\cite{brazil2019pedestrian} exploited sequential labeling policy in the region proposal network to gradually filter out better proposals.
The work of~\cite{chen2018person} employed a two-stage pretrained person detector (Faster R-CNN) and an instance segmentation model for person re-identification. Each detected person is cropped out from the original image and fed to another network.
Wang et al.~\cite{wang2018repulsion} introduced repulsion losses that prevent a predicted bounding box from shifting to neighboring overlapped objects to counter occlusions. Two-stage detectors need to generate proposals in the first stage, and thus are slow for inference in practice.
One-stage detector GDFL~\cite{lin2018graininess} included semantic segmentation, which guided feature layers to emphasize pedestrian regions. Liu et al.~\cite{liu2018learning} extended the single-stage architecture with an asymptotic localization fitting module storing multiple predictors to evolve default anchor boxes. This improves the quality of positive samples while enabling hard negative mining with increased thresholds. 
Similar to pedestrian detection, vehicle detection in ITS also is a popular and challenging computer vision task~\cite{zhao2019object}. 
Current generic vehicle detectors are divided into two categories: CNN-based two-stage detectors and CNN-based one-stage detectors. the representative two-stage detectors include Faster R-CNN~\cite{zhang2016faster}, spatial pyramid pooling (SPP)-net~\cite{he2015spatial}, feature pyramid networks (FPN)~\cite{lin2017feature}, and Mask R-CNN~\cite{he2017mask}. And the representative one-stage detectors include YOLO~\cite{redmon2016you}, Single Shot MultiBox Detector (SSD)~\cite{liu2016ssd}, and  deeply
supervised object detectors (DSOD)~\cite{shen2017dsod}.
The two-step framework is a region proposal-based method, giving a coarse scan of the whole image first and then focusing on regions of interest (RoIs). While one-step frameworks are based on global regression/ classification, mapping straightly from image pixels to bounding box coordinates and class probabilities. Based on these two frameworks, most of the works get promising
results in vehicle detection by combining other methods, such as multitask learning~\cite{brahmbhatt2017stuffnet}, multi-scale representation~\cite{bell2016inside}, and context modeling~\cite{kong2016hypernet}.

\subsubsection{Current methods to overcome challenge}

In traffic sign detection, existing traffic sign datasets are limited in terms of the type and severity of challenging conditions. Metadata corresponding to these conditions is unavailable and it is not possible to investigate the effect of a single factor because of the simultaneous changes in numerous conditions. To overcome this, \cite{temel2019traffic} introduced the CURE-TSDReal dataset, based on simulated conditions that correspond to real-world environments.
An end-to-end traffic sign detection framework Feature Aggregation Multipath Network (FAMN) is proposed in \cite{ou2019famn}. It consists of two main structures named Feature Aggregation and Multipath Network structure to solve the problems of small object detection and fine-grained classification in traffic sign detection.

 A vehicle highlight information-assisted neural network for vehicle detection at night is presented in \cite{mo2019highlight}, which included two innovations: establishing the label hierarchy for vehicles based on their highlights and designing a multi-layer fused vehicle highlight information network.
Real-time vehicle detection for nighttime situations is presented in~\cite{bell2021novel}, where images include flashes that occupy large image regions, and the actual shape of vehicles is not well defined. By using a global image descriptor along with a grid of foveal classifiers, vehicle positions are accurately and efficiently estimated.
AugGAN~\cite{lin2020gan} is an unpaired image-to-image translation network for domain adaptation in vehicle detection. It quantitatively surpassed competing methods for achieving higher nighttime vehicle detection accuracy because of better image-object preservation. 
A stepwise domain adaptation (SDA) detection method is proposed to further improve the performance of CycleGAN by minimizing the divergence in cross-domain object detection tasks in~\cite{li2022stepwise}. In the first step, an unpaired image-to-image translator is trained to construct a fake target domain by translating the source images to similar ones in the target domain. In the second step, to further minimize divergence across domains, an adaptive CenterNet is designed to align distributions at the feature level in an adversarial learning manner.

\subsection{Autonomous driving perception: segmentation}
\subsubsection{Models and algorithms}
Image segmentation contains three sub-tasks: semantic segmentation, instance segmentation, and panoptic segmentation.
Semantic segmentation is a fine prediction task to label each pixel of an image with a corresponding object class, instance segmentation is designed to identify and segment pixels that belong to each object instance, while panoptic segmentation unifies semantic segmentation and instance segmentation such that all pixels are given both a class label and an instance ID ~\cite{gu2022review}.

YOLACT~\cite{bolya2019yolact} splits instance segmentation into two parallel sub-architectures. Protonet architecture extracts spatial information by generating a certain number of prototype masks, and Head architecture generates the mask coefficients and object locations. In addition, it employs Fast NMS rather than traditional NMS to reduce post-processing time.
CenterMask~\cite{lee2020centermask} is an efficient anchor-free instance segmentation, that adds a novel spatial attention-guided mask (SAG-Mask) branch to a one-stage object detector in the same vein as Mask R-CNN.
Path Aggregation Network (PANet)~\cite{liu2018path} is proposed to integrate comprehensively low-level location information and high-level semantic information. Based on Feature Pyramid Networks (FPN)~\cite{lin2017feature}, PANet designs a bottom-up context information aggregation structure, which can integrate different levels of features.
A learning-based snake algorithm~\cite{peng2020deep} is proposed for real-time instance segmentation, which introduces the circular convolution for efficient feature learning on the contour and regresses vertex-wise offsets for the contour deformation.
A top-down instance segmentation framework based on explicit shape encoding, named ESE-Seg, is proposed to reduce the computational time of instance segmentation by decoding the multiple object shapes with tensor operations, thus performing the instance segmentation at almost the same speed as the object detection \cite{xu2019explicit}. 
Multi-task Network Cascades (MNC)~\cite{dai2016instance} divide the instance segmentation task into three cascaded sub-tasks: instance discrimination (discriminating different instances with non-semantic bounding boxes), mask generation (generating pixel-level masks), and object classification (assigning a semantic label to each instance). 
Hybrid Task Cascade (HTC) is proposed for instance segmentation in~\cite{chen2019hybrid}. It interweaves box and mask branches for joint multi-stage processing, adopts a semantic segmentation branch to provide spatial context, and integrates complementary features together in each stage.

\subsubsection{Current methods to overcome challenge}

Recent directions in segmentation include weakly-supervised semantic segmentation~\cite{wang2020self, sun2020mining}, domain adaptation~\cite{chen2019domain,liu2021source}, multi-modal data fusion~\cite{feng2020deep,cortinhal2021semantics}, and real-time semantic segmentation~\cite{yu2018bisenet,nirkin2021hyperseg,gao2021mscfnet}.

TS-Yolo~\cite{wan2021novel} is a CNN-based model for accurate traffic detection under severe weather conditions using new samples from data augmentation. The data augmentation was conducted using copy-paste strategy, and a large number of new samples were constructed from existing traffic-sign instances. Based on YoloV5, MixConv was also used to mix different kernel sizes in a single convolution operation so that patterns with various resolutions can be captured. 
Detecting and classifying real-life small traffic signs from large input images is difficult due to them occupying fewer pixels relative to larger targets. To address this, Dense-RefineDet~\cite{sun2020dense} applies a single-shot, object-detection framework to maintain a suitable accuracy–speed trade-off.
An end-to-end traffic sign detection framework Feature Aggregation Multipath Network is proposed in \cite{ou2019famn} to solve the problems of small object detection and fine-grained classification in traffic sign detection.

\subsection{Cooperative perception}

\subsubsection{Models and algorithms}

In connected autonomous vehicles (CAV), cooperative perception can be performed at three levels depending on the type of data: early fusion (raw data), intermediate fusion (preprocessed data), where intermediate neural features are extracted and transmitted, and late fusion (processed data), where detection outputs (3D bounding box position, confidence score) are shared. Cooperative perception studies how to leverage visual cues from neighboring connected vehicles and infrastructure to boost the overall perception performance~\cite{xu2022v2x}.

1) Early fusion: \cite{chen2019cooper} fuses the sensor data collected from different positions and angles of connected vehicles using raw-data level LiDAR 3D point clouds, and a point cloud-based 3D object detection method is proposed to work on a diversity of aligned point clouds. DiscoNet~\cite{li2021learning} leverages knowledge distillation to enhance training by constraining the corresponding features to the ones from the network for early fusion.

2) Intermediate fusion: F-Cooper~\cite{chen2019f} provides both a new framework for applications On-Edge, servicing autonomous vehicles as well as new strategies for 3D fusion detection.
Wang, et al.~\cite{wang2020v2vnet} proposed a vehicle-to-vehicle (V2V) approach for perception and prediction that transmits compressed intermediate representations of the P\&P neural network.
Xu, et al.~\cite{xu2021opv2v} proposed an Attentive Intermediate Fusion pipeline to better capture interactions between connected agents within the network. 
A robust cooperative perception framework with vehicle-to-everything (V2X) communication using a novel vision Transformer is presented in \cite{xu2022v2x}.

3) Late fusion: Car2X-based perception~\cite{rauch2012car2x} is modeled as a virtual sensor in order to integrate it into a high-level sensor data fusion architecture.

\subsubsection{Current methods to overcome challenge}
To reduce the communications load and overhead, an improved algorithm for message generation rules in collective perception is proposed~\cite{thandavarayan2020generation}, which improves the reliability of V2X communications by reorganizing the transmission and content of collective perception messages.
This paper~\cite{yoon2021performance} presents and evaluates a unified cooperative perception framework containing a decentralized data association and fusion process that is scalable with respect to participation variances. The evaluation considers the effects of communication losses in the ad-hoc V2V network and the random vehicle motions in traffic by adopting existing models along with a simplified algorithm for individual vehicle’s on-board sensor field of view.
AICP is proposed in~\cite{zhou2022aicp}, the first solution that focuses on optimizing informativeness for pervasive cooperative perception systems with efficient filtering at both the network and application layers. To facilitate system networking, they also use a networking protocol stack that includes a dedicated data structure and a lightweight routing protocol specifically for informativeness-focused applications.

\subsection{Vehicle interaction}

\subsubsection{Models and algorithms}
 Computer vision methods can be used to detect and classify crash and near-crash events based on motion and trajectories.
CNN, in the form of modified YOLOv3, is used to detect objects and extract semantic information about the road scene from onboard camera data from the SHRP2 dataset in \cite{Taccari2018crash}. Optical flow is calculated from consecutive frames to track objects and to generate features (hard deceleration, the maximum area of the largest vehicle, time to collision, etc.) that are combined with telematics data (speed and 3-axis acceleration) to train a random forest classifier on safe, near-crash, and crash events. 
 
 Dashcam videos were used to train a Dynamic Spatial Assistance (DSA) network to distribute attention to objects and model temporal dependencies in \cite{chan2016anticipating}. The method was able to predict crashes around 2 seconds in advance. 
 Understanding multi-vehicle interaction in urban environments is challenging, and model-based methods may require prior knowledge, so a more general approach is explored in \cite{Zhang2019generalInteraction}, where YOLOv3 is used for object tracking from traffic camera video from the NGSIM dataset and a Gaussian velocity field is used to describe the interaction behaviors between multiple vehicles. From this, an 11-layer deep autoencoder learns the latent low dimensional representations for each frame, followed by a hidden semi-Markov model with a hierarchical Dirichlet process, which optimizes the number of interaction patterns, to cluster representations into traffic primitives corresponding to the interaction patterns. The pipeline can be used to analyze complex multi-agent interactions from traffic video.
 DL methods are able to extract semantic descriptions from video, like in \cite{YuLiCrash2021}, which can give advance warning of risky situations. A scenario-wise spatio-temporal attention guidance system was created by data mining from descriptive semantic variables in fatal crash data to support the design of a model based on YOLOv3 for evaluating crash risk from dashcam footage. The attention guidance extracted semantic descriptions like ``pedestrian'', ``school bus'' and ``atmospheric condition'', followed by DL to optimize attention on these variables to identify clusters and associate scene features with crash features.

 \subsubsection{Current methods to overcome challenges}
 While most vehicle interaction methods reviewed thus far make little mention of the practical challenges in variable weather and lighting, \cite{ZhangGrade2018} highlights a background learning method specifically to adapt to changing lighting conditions and headlight illumination in surveillance footage and even utilizes a threshold based noise removal for rainy conditions to detect near-miss events at grade crossings.
 Domain adaptation, an example of transfer learning, was employed in \cite{li2021domain} to make use of labeled daytime footage for vehicle detection in unlabeled nighttime images by a generative adversarial network called CycleGAN \cite{zhu2017unpaired}, which can be used with many real-world deep learning computer vision applications.
YouTube dashcam footage was used for crash detection in an ensemble multimodal DL method, based on the gated recurrent unit (GRU) and CNN, that uses both video and audio data \cite{CHOI2021crashDet}. The real-world data consists of positive clips containing crashes and negative clips containing normal driving. A crowd-sourced dashcam video dataset was also contributed by \cite{chan2016anticipating} for accident anticipation containing scenarios like crowded streets, complicated road environments, and diversity of accidents.
To address low-visibility conditions like rain, fog, and nighttime footage, \cite{wang2020vision} used Retinex image enhancement algorithm for preprocessing and YOLOv3 for object detection, followed by a decision tree to classify crashes. It balances dynamic range and enhances edges, but congested mixed-flow traffic, lower-quality video, and fast vehicles are still major sources of error. 
The use of deep convolutional autoencoders for representation learning complemented with vehicle tracking is used to detect accidents from surveillance footage in \cite{singh2019accident}. The testing was performed on data collected during bright sunlight, night, early morning, and also from a variety of cameras and angles. However, there are significant false alarms caused by low visibility, occlusions, and large variations in traffic patterns.  
The lack of near-miss data can be met with combining vehicle event recorder data and object detection from an onboard camera as proposed in \cite{YAMAMOTO2022nearmiss}. By extracting two deep feature representations that consider the car status and the surrounding objects, the deep learning method can label near-miss events. While the method does not claim to be real-time, it can generate large volumes of labeled training data for near-crash events.

 A method to detect cycling near-misses from front view video is developed in \cite{ibrahim2021cyclingnet} using optical flow, CNN, LSTM, and a fully connected prediction stage. The method was trained with complex urban environments and also contributes to a large dataset containing labeled near-miss events. A ResNet-based model was used to detect pedestrians and evaluate risk from a near-miss dataset in \cite{suzuki2017pedestrian}. The dataset contains videos from different vehicles, places (intersections, city, major roads), day and night time, and weather conditions. However, the model suffers from overfitting as a result of having only near-miss data for training. 

\subsection{Road user behavior prediction}

\subsubsection{Models and algorithms}
Trajectory prediction from videos is useful for autonomous driving, traffic forecasting, and congestion management. Older works in this domain focused on homogeneous agents such as cars on a highway or pedestrians in a crowd, whereas heterogeneous agents were only considered in sparse scenarios with certain assumptions like lane-based driving.
A long short-term memory (LSTM) and CNN hybrid network, that learns the relationship between pairs of heterogeneous agents, was developed in \cite{traphic2019} to extract agent shape, velocity, and traffic concentration, which are passed through LSTMs to generate horizon and neighborhood maps, which then go through convolution networks to produce latent representations that are passed through a final LSTM to predict the trajectory. It can perform accurately in dense, heterogeneous, urban traffic conditions in real time. The paper also contributes a new labeled dataset captured from crowded Asian cities.
In order to be useful, trajectory prediction needs to take into account the motion of surrounding objects and inter-object interactions in real time. Therefore, a different approach to motion prediction is discussed in \cite{grip2019} based on the graph convolutional model, which takes trajectory data as input and represents the interactions of nearby objects and extracts features. The graph model output is then passed into an encoder-decoder LSTM model for robust predictions that can consider the interaction between vehicles. The method enables 30\% higher prediction accuracy in addition to 5x faster execution. The algorithm uses trajectory data that has already been extracted from surveillance video data like NGSIM \cite{ngsim2007us}. 
 In \cite{TRIPICCHIO2022trajectory}, vehicle trajectory of vehicles is calculated using Lucas-Kanade algorithm on dashcam video. Synthetic data was also used for augmenting the dataset to train an LSTM network to predict future motion and an SVM is used to classify the action, for eg. changing lanes. The method predicts the next 6 seconds of motion on highways with 92\% accuracy.

 \subsubsection{Current methods to overcome challenges}
 
  The dynamics of vulnerable road users are described by a Switching Linear Dynamical System (SLDS) in \cite{kooij2019context} and extended with a dynamic bayesian network using context from features extracted from vehicle-mounted stereo cameras focusing on both static and dynamic cues. The approach can work in real-time, providing accurate predictions of road user trajectories. It can be improved by the inclusion of more context like traffic lights and pedestrian crossings.
 The use of onboard camera and LiDAR along with V2V communication is explored in \cite{choi2021trajectory} to predict trajectories using the random forest and LSTM architecture. YOLO is used to detect cars and provide bounding boxes, while LiDAR provides subtle changes in position, and V2V communication transmits raw values like steering angles to reduce the uncertainty and latency of predictions. 
 
  The TRAF dataset was used in \cite{chandra2019trajectory} for robust end-to-end real-time trajectory prediction from still or moving cameras. Mask R-CNN and reciprocal velocity obstacles algorithm are used for multi-vehicle tracking. The last 3 seconds of tracking are used to predict the next 5 seconds of trajectory as in \cite{traphic2019}, with the added advantage of being end-to-end trainable and not requiring annotated trajectory data. The paper also contributes TrackNPred, a python-based library that contains implementations of different trajectory prediction methods. It is a common interface for many trajectory prediction approaches and can be used for performance comparisons using standard error measurement metrics on real-world dense and heterogeneous traffic datasets.

 Most DL methods for trajectory prediction do not uncover the underlying reward function, instead, they only rely on previously seen examples, which hinders generalizability and limits their scope. In \cite{fernando2020irl}, inverse reinforcement learning is used to find the reward function so that the model can be said to have a tangible goal, allowing it to be deployed in any environment.  
 Transformer-based motion prediction is performed in \cite{liu2021multimodal} to achieve state-of-the-art multimodal trajectory prediction in the Agroverse dataset. The network models both the road geometry and interactions between the vehicles. 
 Pedestrian intention in complex urban scenarios is predicted by graph convolution networks on spatio-temporal graphs in \cite{liu2020pedestrian}. The method considers the relationship between pedestrians waiting to cross and the movement of vehicles. While achieving 80\% accuracy on multiple datasets, it predicts intent to cross one second in advance. On the other hand, pedestrians modeled as automatons, combined with SVM without the need for pose information, result in longer predictions but lack the consideration of contextual information \cite{jaya2020pedestrian}.

\subsection{Traffic anomaly detection}
\subsubsection{Models and Algorithms}

Traffic surveillance cameras can be used to automatically detect traffic anomalies like stopped vehicles and queues. The detection of low-level image features like corners of vehicles has been used by \cite{Albiol2011} to demonstrate queue detection and queue length estimation without object tracking or background removal in different lighting conditions. Tracking methods based on optical flow can not only provide queue length, but also speed, vehicle count, waiting time, and time headway. In \cite{shiraziQueue2015}, the authors use optical flow assuming constant short-term brightness to detect vehicle features and successfully track them even with occlusions. The speed of individual vehicles can be estimated, allowing the detection of stopped vehicles or queue formation. Trajectory analysis has also been used to identify illegal or dangerous movements \cite{nowosielski2016traj}. The background subtraction-based approaches are, however, limited to favorable scenarios and do not generalize well. On the other hand, the system has been physically deployed to analyze surveillance camera video.

An interesting method is applied in \cite{LiAnomaly2016} involving partitioning the video into spatial and temporal blocks, local invariant features are then learned from traffic footage to create a visual codebook of the image descriptors using Locality-constrained Linear Coding. Then, a Gaussian distribution model is trained to learn the probabilities corresponding to normal traffic, which can be used to detect anomalies. The image description makes it more robust to lighting, perspective, and occlusions.
Armstrong, et al. \cite{armstrongCVPR2021} proposed a decision tree-based DL approach for anomaly detection using YOLOv5 for vehicle detection, followed by background estimation, then a decision tree considers factors like vehicle size, likelihood, and road feature mask to eliminate false positives. Adaptive thresholding allows for robustness under variable illumination and weather conditions. A perspective map approach is discussed by \cite{bai2019cvpr}, which models the background using road segmentation based on a traffic flow frequency map, then the perspective is detected from linear regression of object sizes based on ResNet50. Finally, a spatial-temporal matrix discriminating module does thresholding on consecutive frames to detect anomalous states.

\subsubsection{Current methods to overcome challenges}
 
Anomaly detection relies on surveillance cameras which usually provide a view far along the road, but vehicles in the distance occupy only a few pixels which make detection difficult. Thus, \cite{li2020multi} uses pixel-level tracking in addition to box-level tracking for multi-granularity. The key idea is mask extraction based on frame difference and vehicle trajectory tracking based on Gaussian Mixture Model to eliminate moving vehicles combined with segmentation based on frame changes to also eliminate parking zones. Anomaly fusion uses the box and pixel-level tracking features with backtracking optimization to refine predictions. 
 Surveillance cameras are prone to shaking in the wind, so video stabilization preprocessing was performed before using two-stage vehicle detection in the form of Faster R-CNN and Cascade R-CNN \cite{zhao2021good}. An efficient real-time method for anomaly detection from surveillance video decouples the appearance and motion learning into two parts \cite{Li2021anomaly}. First, an autoencoder learns appearance features, then 3D convolutional layers can use latent codes from multiple past frames to predict features for future frames. A significant difference between predicted and actual features indicates an anomaly. The model can be deployed on edge nodes near the traffic cameras, and the latent features appear to be robust to illumination and weather changes compared to pixel-wise methods.
 
 To shed reliance on annotated data for anomalies, an unsupervised one-class approach in \cite{pawar2021anomaly} applies spatio-temporal convolutional autoencoder to get latent features, stacks them together, and a sequence-to-sequence LSTM learns the temporal patterns. The method performs well on multiple real-world surveillance footage datasets, but not better than supervised training methods. The advantage is that it can be indefinitely trained on normal traffic data without any labeled anomalies.

\begin{table*}[!ht]
 \centering
  \renewcommand\arraystretch{1.25}
  \caption{Computer Vision Applications in Transportation}
\begin{tabularx}{\textwidth}{|X|X|X|}\hline
 \textbf{Application} & \textbf{Methods} & \textbf{Main Challenges} \\  \hline
 Traffic flow parameter estimation & YOLOv3 + DeepSORT on surveillance videos \cite{hong2020traffic}, YOLOv2 + SPP \cite{kim2019multi}, AlexNet + optical flow + Gaussian mixture model \cite{ke2018multi}, Faster R-CNN + SORT \cite{fedorov2019traffic} & Single metric for performance evaluation, weather and illumination, realtime video processing, diversity of surveillance camera view angles \\  \hline
 Congestion detection and prediction & AlexNet + YOLO \cite{chakraborty2018traffic}, AlexNet + VGGNet \cite{wang2020towards}, YOLO + Mask R-CNN \cite{impedovo2019vehicular}, YOLOv3 \cite{rashmi2020vehicle}, YOLOv4 \cite{sonnleitner2020traffic}, Faster R-CNN \cite{gao2021novel} & Multi-lane highways with exits or barriers, parked vehicles, lack of lane markings, weather and illumination, realtime processing, variety of surveillance camera angles \\  \hline
 Autonomous driving perception: detection & Sparse R-CNN \cite{cao2021traffic}, YOLOv3 \cite{wan2021efficient}, SegNet + U-Net \cite{kamal2019automatic}, Faster R-CNN + Inception-Resnet-v2 \cite{kim2018efficient}, CNN + ConvLSTM \cite{zou2019robust}, GAN \cite{ghafoorian2018gan} & Weather, shadows, complex road environment, occlusion, small objects, degraded lane markings, motion blur  \\  \hline
 Autonomous driving perception: segmentation & YOLACT \cite{bolya2019yolact}, CenterMask \cite{lee2020centermask}, Path Aggregation Network \cite{liu2018path}, ESE-Seg \cite{xu2019explicit}  & Accuracy vs. processing speed trade-off, occlusion, illumination, shadows, weather, lack of generalizability  \\  \hline
 Cooperative perception & Early fusion \cite{chen2019cooper}, DiscoNet \cite{li2021learning}, Intermediate fusion F-Cooper \cite{chen2019f}, V2V intermediate fusion \cite{wang2020v2vnet}, Attentive intermediate fusion \cite{xu2021opv2v}, V2X + Vision transformer \cite{xu2022v2x}, Late fusion \cite{rauch2012car2x}  & Delay, packet loss, calibration, synchronization, non-overlapping points of view, computational cost \\  \hline
 Vehicle interaction & YOLOv3 + optical flow \cite{Taccari2018crash}, Dynamic Spatial Assistance (DSA) RNN \cite{chan2016anticipating}, YOLOv3 + Gaussian velocity field + hidden semi-Markov \cite{Zhang2019generalInteraction}, spatio-temporal attention guidance + YLOv3 \cite{YuLiCrash2021}, Domain adaptation + GAN  \cite{li2021domain}, Gated recurrent unit + CNN \cite{CHOI2021crashDet}, Retinex + YOLOv3 \cite{wang2020vision}, optical flow + CNN + LSTM \cite{ibrahim2021cyclingnet}, ResNet \cite{suzuki2017pedestrian}  & Lack of near-crash data or other specific situations, underrepresentation of bicycles, complex urban traffic, weather, illumination, occlusion  \\  \hline
 Road user behavior prediction & LSTM + CNN \cite{traphic2019}, LSTM + SVM \cite{TRIPICCHIO2022trajectory}, Switching Linear Dynamical System + bayesian network \cite{kooij2019context}, YOLO + LSTM + random forest \cite{choi2021trajectory}, Transformer \cite{liu2021multimodal}  & Lack of benchmark to evaluate methods, realtime computation, faulty sensors, generalizability to different driving scenarios, occlusion \\  \hline
 Traffic anomaly detection & Gaussian mixture model + Optical flow \cite{shiraziQueue2015}, GMM background modeling + AdaBoost classifier + Modified Hausdorff Distance \cite{nowosielski2016traj}, Locality-constrained Linear Coding + Gaussian distribution model \cite{LiAnomaly2016}, YOLOv5 + decision tree + adaptive thresholding \cite{armstrongCVPR2021}, ResNet50 + spatial-temporal matrix discriminating module \cite{bai2019cvpr}, spatio-temporal convolutional autoencoder + LSTM \cite{pawar2021anomaly}  & Vehicles far from camera, camera angles, illumination, lack of generalizability between sparse and dense conditions, camera shaking by wind \\  \hline
 Edge computing & Integer linear programming +  fast heuristics \cite{CuiEdge2020}, SSD + SORT \cite{ke2020nearcrash}, Deep Deterministic Policy Gradient + V2V networking \cite{DaiEdge2019}, cloud-edge hybrid + Global foreground modeling + Gaussian mixture model \cite{liu2021edge}, YOLOv3 \cite{wan2022edge}, federated learning \cite{federated2019}, spectral clustering compression \cite{chen2021iov} & Power consumption, heterogeneous data sources, cybersecurity, wireless noise, scalability, neural network pruning and model compression, installation and maintenance costs   \\  \hline
 \end{tabularx}
 \label{table1}
\end{table*}

\subsection{Edge computing}

\subsubsection{Models and algorithms}
 
 Computer vision in ITS requires efficient infrastructure architecture to analyze data in real time. If all acquired video streams are sent to a single server, the required bandwidth and computation would not be able to provide a usable service. For example, edge computing architecture for real-time automatic failure detection using a video usefulness metric was explored in \cite{sun2020edge}. Only video deemed to be useful is transmitted to the server, while malfunction of the surveillance camera, or obstruction of view, is automatically reported. Edge-cloud-based computing can implement DL models, not just for computer vision tasks, but also for resource allocation and efficiency \cite{XieDLCV2021}. Passive surveillance has now been superseded in literature by the increasing availability of sensor-equipped vehicles that can perform perception and mapping cooperatively \cite{zhangEdge2020}.
 
 Onboard computing resources in vehicles are often not powerful enough to process all sensor data in real time, and applications like localization and mapping can be very computationally intensive. The internet of things (IoT) architecture allows for edge nodes to offload that computation and provide results at low latency to nearby users \cite{ferdowsi2019edge}. This approach can avoid multiple cars doing the same computation with similar inputs. 
 One technique to offload computation tasks is discussed in \cite{CuiEdge2020}, combining integer linear programming for offline scheduling optimization and heuristics for online, real-world deployment. The authors compress 3D point cloud LIDAR data collected from the vehicle's sensor and send it to the edge node for classification and feature extraction. 
 A deep reinforcement learning algorithm known as Deep Deterministic Policy Gradient is proposed in \cite{DaiEdge2019}, which can dynamically allocate computing and caching resources throughout the network. Future work in this direction will handle multiple communication channels, interference management, forecast handover, and bandwidth allocation. In the macro scale, V2V communication can be used for traffic parameter estimation and management with sparse connectivity, while higher connected vehicle market penetration will allow safety applications like collision avoidance \cite{Dey2016v2v}. 
 
 Applications for vehicles can include near-crash detection, navigation, video streaming, and smart traffic lights. The onboard unit can also be used as a mobile cache, and communicate with other vehicles via V2V networking. Real-time near-crash detection using edge computing was developed in \cite{ke2020nearcrash}. The system uses dashcam video for SSD vehicle and pedestrian detection, followed by SORT for tracking to estimate the time to collision (TTC). It was tested on online datasets and on real cars and buses. The detected events, along with CAN bus messages was used to filter irrelevant data, saving bandwidth for data collection. A practical deployment of parking surveillance using edge-cloud computing was presented in \cite{rke2021parking}, the edge device performs detection and transmits the bounding box and object types to the server, which uses this information for labeling and tracking. A different approach by \cite{bura2018parking} focused on vehicle tracking from top view cameras and number plate recognition from ground-level cameras for real-time occupancy information and to automatically charge a vehicle for the time it was parked.
 Large-scale traffic monitoring using computer vision and edge computing was detailed in \cite{liu2021edge} where edge nodes close to surveillance cameras can process low-resolution videos to monitor traffic, detect congestion, and detect speed if the available bandwidth is low. If high bandwidth to the server is available, high-quality video will be sent for similar processing. Edge computing for vehicle detection is examined in \cite{wan2022edge}. The algorithm divides the traffic video into segments of interest and then uses YOLOv3 for vehicle detection in real-time on the edge node, and the extracted clips are used as training data for the edge server.
 \subsubsection{Current methods to overcome challenges}
 
 One problem with large-scale DL is that the huge quantity of data produced cannot be sent to a cloud computer for training. Federated learning \cite{federated2015} has emerged as a solution to this problem, especially considering the heterogeneous data sources, bandwidth, and privacy issues \cite{Zhou2021edge}. Training can be performed on edge nodes or edge servers, with the results being sent to the cloud to aggregate in the shared deep learning model \cite{zhangEdge2020}. Federated learning is also robust to failure of individual edge nodes \cite{federated2019}.
 Concerns of bandwidth, data privacy, and power requirements are addressed in \cite{song2018edge} by transferring only inferred data from edge nodes to the cloud, in the form of incremental and unsupervised learning. In general, the processing of data on the edge to reduce bandwidth has the pleasant side effect of anonymizing the transmitted data \cite{bart2019edge}. Another effort to reduce bandwidth requirements employs spectral clustering compression performed on spatio-temporal features needed for traffic flow prediction \cite{chen2021iov}. 
 
 Deep learning models cannot be directly exported to mobile edge nodes, as they are usually too computationally intensive. Neural network pruning both in terms of storage and computation was introduced in \cite{han2015pruning}, while implementation of the resulting sparse network on hardware is discussed in \cite{sparseNN2018}, achieving multiple orders of magnitude increase in efficiency. A general lightweight CNN model was developed for mobile edge units in \cite{zhou2019lightweight}, matching or outperforming AlexNet and VGG-16 while being a fraction of the size and computation cost. Edge computing-based traffic flow detection using deep learning was deployed by \cite{chen2020edge} where YOLOv3 was trained and pruned, along with DeepSORT, to be deployed on the edge device for real-time performance.
 A thorough review of deploying compact DNNs on low-power edge computers for IoT applications can be found in \cite{zhang2021compact}. They note that the diversity and quantity of DNN applications require an automated method for model compression beyond traditional pruning techniques.

\section{Future Directions}\label{Sec:Future}


\subsection{For solving data challenges}

While a large quantity of data is essential for training deep learning models, often the quality is the limiting factor in training performance. Data curation is a necessary process to include edge cases and train the model on representative data from the real world. Labeling vision data, especially in complex urban environments is a labor-intensive task performed by humans. It can be sped up by first using existing object detection or segmentation algorithms based on the relevant task to automatically label the data. Then this can be further checked by humans to eliminate errors by the machine, thus creating a useful labeled dataset. There is also a need for datasets that include multiple sensors from different views for training cooperative perception algorithms. Collecting such data is bound to be challenging because of hardware requirements and synchronization issues but it is possible to do with connected vehicles and instrumented intersections similar to the configuration that will be deployed. 

The problems associated with poor quality or viewing angle of real-world cameras can be mitigated by using realistic CCTV benchmarks and datasets that include a wide variety of surveillance footage, including synthetic video \cite{revaud2021ICCV}. Data-driven simulators like \cite{amini2021vista} use high-fidelity datasets to simulate cameras and LiDAR, which can be used to train DL models with data that is hard to capture in the real world \cite{azfar2022efficient}. Such an approach has shown promise in end-to-end reinforcement learning of autonomous vehicle control \cite{amini2020simulation}. Domain adaptation techniques are expected to be further extended to utilize synthetic data and conveniently collected data. 

Sub-fields in transfer learning, especially few-shot learning and zero-shot learning, will be extensively applied with expert knowledge to address the lack of data challenges, such as corner case recognition in ITS and AD. Likewise, new unsupervised learning and semi-supervised learning models are expected in the general field of real-world computer vision. Future work in vision transformer explainability will allow for more comprehensive insights based on aggregated metrics over multiple samples \cite{aflalo2022vl}. Interpretability research is also expected to evaluate differences between model-based and model-free reinforcement learning approaches \cite{atakishiyev2021expl}.

Data decentralization is a well-recognized trend in ITS. To address issues like data privacy, large-scale data processing, and efficiency, crowdsensing \cite{ning2021blockchain} and federated learning on vision tasks \cite{liu2020fedvision} are unavoidable future directions in ITS and AD. Additionally, instead of the traditional way of training a single model for a single task, multiple downstream tasks learning with a generalized foundation model, e.g., Florence \cite{yuan2021florence}, is a promising trend to deal with various data challenges. Another mechanism is data processing parallelism in ITS coupled with edge computing for multi-task (e.g., traffic surveillance and road surveillance) learning \cite{ke2022real}. 

\subsection{For solving model challenges}

Deep learning models are trained until they achieve good accuracy, but real-world testing often reveals weaknesses in edge cases and complex environmental conditions. There is a need for online learning for such models to continue to improve and adapt to real-world scenarios otherwise they cannot be of practical use. If online training is not possible due to a lack of live feedback on the correctness of the predictions, the performance must be analyzed periodically with real data stored and labeled by humans. This can serve as a sort of iterative feedback loop, where the model does not need to be significantly changed, just incrementally retrained based on the inputs it finds most challenging. One possible way to partially automate this would be to have multiple different redundant architectures using the same input data to make predictions along with confidence scores. If the outputs do not agree, or if the confidence scores are low for a certain output, that data point can be manually labeled and added to the training set for the next training iteration. 

Complex deep learning models deployed to edge devices need to be more efficient through methods such as pruning \cite{han2015pruning}. Simple pruning methods can improve CNN performance by over 30\% \cite{li2016pruning}. Depending on the specific architecture, the models may also be split into different functional blocks deployed on separate edge units to minimize bandwidth and computation time \cite{sufian2021deep}. A foreseeable future stage of edge AI is "model training and inference both on the edge," without the participation of cloud datacenters.

In recent years much work has been done towards explainable AI, especially in computer vision. CNNs have been approached with three explainability methods: gradient-based saliency maps, Class Activation Mapping, and Excitation Backpropagation \cite{zhang2018top}. These methods were extended for graph convolutional networks in \cite{pope2019explainability}, pointing out patterns in the input that correspond with the classification. Generic solutions for explainability have been presented in \cite{chefer2021generic} for both self-attention and co-attention transformer networks. While it is not straightforward to apply these methods to transportation applications, some efforts have been made to understand deep spatio-temporal neural networks dealing with video object segmentation and action recognition quantifying the static and dynamic information in the network and giving insight into the models and highlighting biases learned from datasets \cite{kowal2022deeper}.

Cooperative sensing model development is a necessary future direction for better perception in 3D, in order to mitigate the effects of occlusion, noise, and sensor faults. V2X networks and vision transformers have been used for robust cooperative perception, which can support sensing in connected autonomous vehicle platforms \cite{xu2021opv2v}\cite{xu2022v2x}.
Connected autonomous vehicles will also host other deep learning models that can learn from new data in a distributed manner. Consensus-driven distributed perception is expected to make use of future network technologies like 6G V2X, resulting in low latency model training that can enable true level 5 autonomous vehicles \cite{barbieri2022decentralized}.

\subsection{For solving complex traffic environment challenges}

Multimodal sensing and cooperative perception are necessary future avenues of practical research. Different modalities like video, LiDAR, and audio can be used in combinations to improve the performance of methods purely based on vision. Audio is especially useful for detecting anomalies earlier among pedestrians like fights or commotions, and for vehicles in crowded intersections where the visual chaos may not immediately reveal problems like mechanical faults, or minor accidents. Cooperative perception will allow multiple sensor views of the same environment from different vehicles to build a common picture that contains more information than any single agent can perceive thus solving problems of occlusion and illumination. 

There is an increasing trend of using transfer learning to improve model performance in real-world tasks. Initially training the model on synthetic data and fine-tuning with task-specific data reduces the reliability on complex, single use deep learning models and improve real-world performance by retraining on challenging urban scenarios. As aforementioned, domain adaptation, zero-shot learning, few-shot learning, and foundation models are expected transfer learning areas that serve this purpose.

The results of unsupervised methods like in \cite{pawar2021anomaly} can be further improved by online learning in crowded and challenging scenarios after deployment on embedded hardware, as there is an unlimited supply of unlabeled data.  
The lack of theoretical performance analysis regarding the upper bound on false alarm rate in complex environments is discussed as an important aspect of deep learning methods for anomaly detection in \cite{doshi2021pedestrian}. Future research is recommended to include this analysis as well.
It is hard to imagine complete reliance on surveillance cameras for robust, widespread, and economical traffic anomaly detection. The method in \cite{PARSA2020xgboost} includes traffic, network, demographic, land use, and weather data sources to detect traffic. Such ideas can be used in tandem with computer vision applications for better overall performance.
 
Future directions in the application of edge computing in ITS will consider multi-source data fusion along with online learning \cite{xie2021DLITS}. Many factors like unseen shapes of vehicles, new surrounding environments, variable traffic density, and rare events can be too challenging for DL models \cite{ferdowsi2019edge}. This new data could be used for online training of the system. Traditional applications can be extended using edge computing and IoV/IoT frameworks.
Vehicle re-identification from video is emerging as the most robust solution to occlusion \cite{zhao2021CVPR}. However, the inclusion of more spatio-temporal information for learning leads to greater memory and computational usage. Tracklets from one camera view can be matched with other views at different points in time using known features. Instead of using a fixed window, adaptive feature aggregation based on similarity and quality, can be generalized to many multi-object tracking tasks \cite{qian2020adaptive}.
 
Transformers are particularly useful at learning dynamic interactions between heterogenous agents which will be particularly useful in crowded urban environments for detection and trajectory prediction. They can also be used for the detection of anomalies and prediction of potentially hazardous situations like collisions in a multi-user heterogeneous scenario.

\section{Conclusions}\label{Sec:Conclusions}
In real-world scenarios, most of the DL computer vision methods suffer from severe performance degradation when facing different challenges. In this paper, we review the specific challenges for data, models, and complex environments in ITS and autonomous driving. Many related deep learning-based computer vision methods are reviewed, summarized, compared, and discussed. Furthermore, a number of representative deep learning-based applications of ITS and autonomous driving are summarized and analyzed. Based on our analysis and review, several potential future research directions are provided. We expect that this paper could provide useful research insights and inspire more progress in the community.


\footnotesize
\bibliographystyle{IEEEtran}
\bibliography{refs}

\end{document}